# BSH FOR COLLISION DETECTION IN POINT CLOUD MODELS

Mauro Figueiredo[1], João Pereira[2], João Oliveira[3], Bruno Araújo[2],

[1] CIMA-Centro de Investigação Marinha e Ambiental, Instituto Superior de Engenharia, Universidade do Algarve, Faro, Portugal, mfiguei@ualg.pt

[2] IST/INESC-ID, Rua Alves Redol, 9, 1000-029 Lisboa, {bruno.araujo, jap}@inesc-id.pt

[3] C3i - Centro Interdisciplinar de Investigação e Inovação, Instituto Politécnico de Portalegre, Portugal, jfoliveira@estgp.pt

## Abstract

Point cloud models are a common shape representation for several reasons. Three-dimensional scanning devices are widely used nowadays and points are an attractive primitive for rendering complex geometry. Nevertheless, there is not much literature on collision detection for point cloud models.

This paper presents a novel collision detection algorithm for large point cloud models using voxels, octrees and bounding spheres hierarchies (BSH). The scene graph is divided in voxels. The objects of each voxel are organized into an octree. Due to the high number of points in the scene, each non-empty cell of the octree is organized in a bounding sphere hierarchy, based on an R-tree hierarchy like structure. The BSH hierarchies are used to group neighboring points and filter out very quickly parts of objects that do not interact with other models.

Points derived from laser scanned data typically are not segmented and can have arbitrary spatial resolution thus introducing computational and modeling issues. We address these issues and our results show that the proposed collision detection algorithm effectively finds intersections between point cloud models since it is able to reduce the number of bounding volume checks and updates.

## 1. INTRODUCTION

Point cloud models are an increasingly attractive representation used as the basis to capture and measure reality rapidly in an increasing number of applications such as environmental surveying, structure analysis and archaeology [1]. In general interactive virtual environments often need very fast collision detection queries to simulate physical behaviour and to allow the user to interact. However, there is practically no literature on determining collisions between two sets of points.

This paper describes a novel collision detection algorithm for point cloud models.

The scene graph is organized into voxels. To speed up the process of finding collisions, each voxel, is partitioned by an octree. The points of each non-empty cell of the octree are organized in a bounding sphere hierarchy (BSH), like an R-tree data structure, defined in its own local coordinate system. The BSH organizes spatially its geometry, grouping neighbouring points.

Results show that the collision detection algorithm for point cloud models uses effectively the bounding sphere hierarchy to find intersections between point cloud models at interactive rates. In addition, unlike CAD objects which typically contain object hierarchies and are already segmented into surface groups, point data sets derived from laser scanned data do not have such information thus presenting computational issues. The octree was used for this purpose to address these issues and it is a solution that adapts to point sets derived from different laser scanners and spatial settings.

The paper is organized as follows. Section 2 presents collision detection approaches for the determination of intersections between polygonal and point cloud models. Section 3 describes the VIZIR project that highlighted the need to develop an efficient collision detection algorithm for point cloud models. Section 4 describes the data structures for the representation of the scene graph. Section 5 presents the bounding sphere hierarchy that is used for to solve the collision detection problem. Section 6 describes the novel collision detection algorithm and results using laser scanned point sets are presented in Section 7. Conclusions and future work are presented in Section 8.

## 2. RELATED WORK

Currently, there are many implementations of collision detection schemes for interactive systems, most of them only support polygonal models. Frequently, they use bounding volume hierarchies (BVH), spatial subdivision methods and more recently use graphics hardware to accelerate collision detection by hardware. There is a lack of collision detection systems for point cloud models.

Bounding volume hierarchies are frequently used to organize the triangles of an object to improve the performance of the collision detection process, by reducing the number of pairs of bounding volume tests. The classic scheme for hierarchical collision detection is a simultaneous recursive traversal of two bounding volumes trees A and B.

Several types of bounding volumes are available. Bounding spheres can be used [2]. SOLID [3], OPCODE [4], Chen [5] and Lu [6] use axis-aligned bounding boxes (AABB). RAPID [7], V-COLLIDE [8], PQP [9], H-COLLIDE [10], use oriented bounding boxes (OBB). QuickCD [11] and Dop-Tree [12] uses k-dops; and Swift++ [13] uses convex hulls (CH). There are also hybrid approaches like QuOSPOs [14] that use a combination of OBBs and k-dops.

The main advantage of SOLID, OPCODE, Box-Tree, Chen [5] and Lu [6] is that AABBs are faster to intersect.

RAPID approximates 3D objects with hierarchies of oriented bounding boxes (OBBs). The main advantage of RAPID is that OBBs are better approximations to triangles reducing effectively the number of intersecting operations.

V-COLLIDE solves the broad-phase of the collision detection using a sweep-and-prune operation to find pairs of objects potentially in contact. It uses RAPID to find in the narrow phase which pairs of objects intersect.

H-COLLIDE is a framework to find collisions for haptic interactions. It uses a hybrid hierarchy of uniform grids and trees of OBBs to exploit frame-to-frame coherence. It was specialized to find collisions between a point probe against 3D objects.

The QuickCD and Dop-Tree implementations build a hierarchy tree of discrete orientation polytopes (k-dops). The main advantage of using discrete orientation polytopes is that k-dops are better approximations to the underlying geometry than AABBs with the advantage of its low cost compared to OBBs.

Swift++ builds a hierarchy of convex hulls and intersection is tested using a modified Lin-Canny closest feature algorithm.

He [14] uses a hybrid approach that combines OBBs and k-dops called QuOSPOs. This approach provides a tight approximation of the original model at each level.

Another class of hierarchical data structures used for collision detection are spatial partitioning representations: regular grids [15, 16, 17, 18], octrees [19, 20], BSP-trees [21] and R-trees [22].

Spatial subdivisions are a recursive partitioning of the embedding space occupied by objects. In general, spatial partitioning structures are used as a secondary representation for the collision detection process.

The main idea behind all space partitioning methods is to exploit spatial coherency. For each object, we check for collision only objects of the neighborhood, eliminating comparisons with those objects that are far away and therefore cannot be colliding.

Zyda [15] uses grids to find overlapping objects in the broad phase. García-Alonso [16] also uses uniform grids to find exact collisions between 3D objects for the narrow phase. Teschner [23] use uniform grid subdivision combined with hashing to reduce storage requirements for collision and self-collision detection of deforming objects that consist of tetrahedrons. Eits [24] also uses a spatial grid inspired by the work of Teschener together with 1D hash table to find collisions between deformable tetrahedral models. A hybrid approach is presented by Gregory [10] using regular grids, where each occupied grid cell stores an OBBs tree of those triangles on that cell.

Hubbard [19] approach for finding collisions in real time is based on a *time-critical* computing algorithm and on octrees of spheres. Kitamura [20] algorithm for collision detection uses an octree for each object. Ganovelli [18] also associate an octree of axis aligned bounding boxes with each object, and keeps this hierarchy efficiently and dynamically updated for deformable objects.

Luque [21] uses semi-adjusting BSP-trees for representing scenes composed of thousands of moving objects.

Figueiredo [22] combines AABBs with R-trees to implement an efficient collision detection algorithm that determines intersecting surfaces.

Various approaches have been recently introduced using existing graphics accelerated boards (GPU) [25, 26, 27, 28] or dedicated hardware [29] to accelerate collision detection by hardware.

Algorithms using graphics hardware use depth and stencil buffer techniques to determine collisions between convex [25] and non-convex [26] objects. CULLIDE [27] is also a GPU based algorithm that uses image-space occlusion queries and OBBs in a hybrid approach to determine intersections between general models with thousands of polygons. MRC [28] deals with large models composed of dozens of millions of polygons by using the representation of a clustered hierarchy of progressive meshes (CHPM) as a LOD hierarchy for a conservative errorbound collision and as a BVH for a GPU-based collision culling algorithm.

These GPU-based algorithms are applicable to both rigid and deformable models since all the computations are made in the image-space. Collision detection methods using GPUs have the disadvantage that they compete with the rendering process, slowing down the overall frame rate. Furthermore, some of these approaches are pure image based reducing their accuracy due to the discrete geometry representation.

All these collision methods have been applied only to polygonal objects. Recently Klein [30] presented a novel approach for collision detection of point clouds. They construct a point hierarchy of bounding volumes to enclose the points at different levels of the hierarchy. Points are stored in the hierarchy leaves. Each node stores a sufficient sample of the points plus a sphere covering of a part of the surface. Given two point cloud hierarchies, two objects are tested for collision by simultaneous traversal. At the leaves, an intersection is determined by estimating the smallest distance.

Kim [31] et. al show the performance benefits of using compression of out-of-core AABBs for collision detection of polygon models that do not fit in main memory, namely they show that the resources of the CPU can be used to compensate the I/O lag of reading uncompressed data structures.

Figueiredo [32] et. al used AABBs where each box centered at a point had respectively the width of the average inter point distance of the data set and twice the average distance, so as to overlap and better approximate a surface collision. However, they noticed that in some areas such as doorways a closer fit would be more desirable to avoid false collisions. It is not immediately clear whether grouping of spheres provides a tighter bounding volume than groups of cubes for arbitrary objects.

## 3. VIZIR

The VIZIR project sets out to develop new visualization and interaction algorithms of massive out-of-core data to build a virtual reality representation of an historical monument in Portugal. Virtual reality has been adopted by the Cultural Heritage field as a perfect, objective simulation of reality to present hyper realistic reconstructions of monuments as true historical representations [33]. The 3D model of study consists of approximately 700 laser scans of the Batalha monastery, ~1 billion points, exceeding 100 GBytes.

Collision detection is an important interaction cue to help user navigation in the virtual world. Unfortunately not much work exists with solutions for collision detection with point clouds.

Before the full complexity of the model can be addressed, an efficient and reliable collision detection solution is needed for point clouds.

For this purpose a simple scenario (Figure 1 and 2) was designed to evaluate different user collisions that can occur whilst navigating and exploring a 3D point cloud model.

In this scenario a subset of the model was chosen that enabled the user's polygonal avatar, which is represented as a point cloud for collision detection purposes, to pass through open doors, walk alongside walls, but is stopped when colliding with the point cloud.

In addition standard collision detection tests were carried out, and collisions with points obtained from CAD models were also tested.

In the next section we present our solution for efficiently detecting collisions with point clouds.

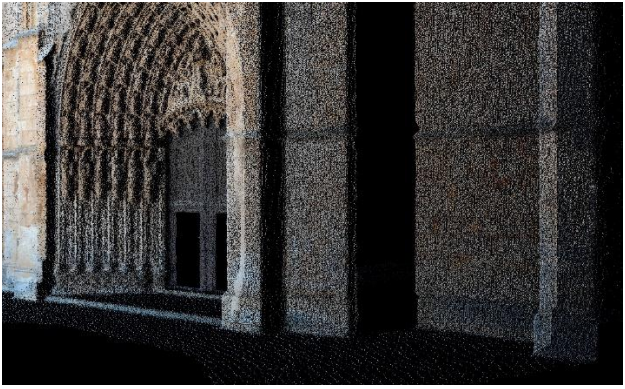

**Figure 1:** First person view of the user in the scenario of interaction whilst navigating the scanned Batalha monastery model.

## 4. POINT CLOUD HIERARCHY

This section presents the data structures that the proposed algorithm uses to find collisions in a large environment where 3D models are described as point clouds.

First a uniform grid that divides the scene graph into $N \times N \times N$ cubic cells of equal volume is used, thus building a grid of voxels. Each voxel is segmented using an octree. Each leaf node of the octree has a list of the objects and the points occupying that region. To study the various user scenarios described in the previous section the voxel containing the entrance to the monastery was used (Figure 2). In a future scenario, each data structure associated with a voxel could be compressed and neighbour voxels to a user's position loaded and uncompressed into a LRU queue [34].

To determine colliding objects at each voxel, the approach presented in this paper uses bounding volume spheres hierarchies, to find collisions between pairs of 3D objects defined as point clouds. Each object is represented by a BSH, R-tree like data structure, in its own local coordinate system. The bounding sphere hierarchy structure is used to filter out portions of the object that cannot intersect.

The choice of bounding volume type influences performance of the collision detection process. The implementation of the collision detection algorithm presented in this paper uses spheres because they use less memory than axis-aligned bounding boxes and are faster to update.

It was decided to use a hierarchy like an R-tree [35] to build bounding sphere hierarchy and organize 3D geometry of objects. R-trees are a good choice for collision detection because first, at any level of the tree, each primitive is associated with only a single node. Secondly, in an R-tree all leaf nodes appear on the same level. Third, because the depth of a R-tree storing *n* primitives is $\log_m n$, *m* is the minimum number of children of a node. And finally, because that the total number of primitives stored in a R-tree equals the number of original primitives.

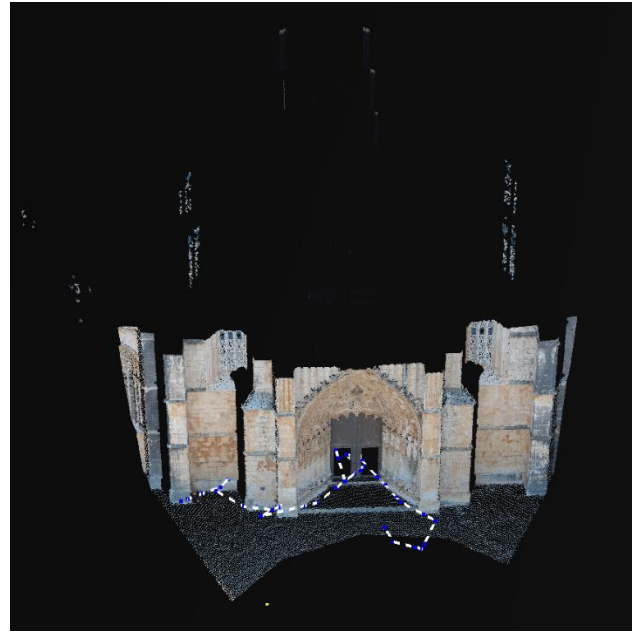

**Figure 2:** Walkthrough collision test scenarios between the Al avatar model comprised of 3617 points (lower left) and the 587923 point cloud belonging to a single voxel.

The points of each voxel are partitioned using an octree. Each partition $P_i$ of the octree is represented by an R-tree data structure in its own local coordinate system (Figure 3) to speed up the process of finding collisions. The R-tree is built, grouping neighbouring points. The leaf nodes are the bounding spheres $S_i$ that enclose the points that define the object. For two objects, it checks for collisions between points which are in the neighbourhood, eliminating comparisons with those that are far away.

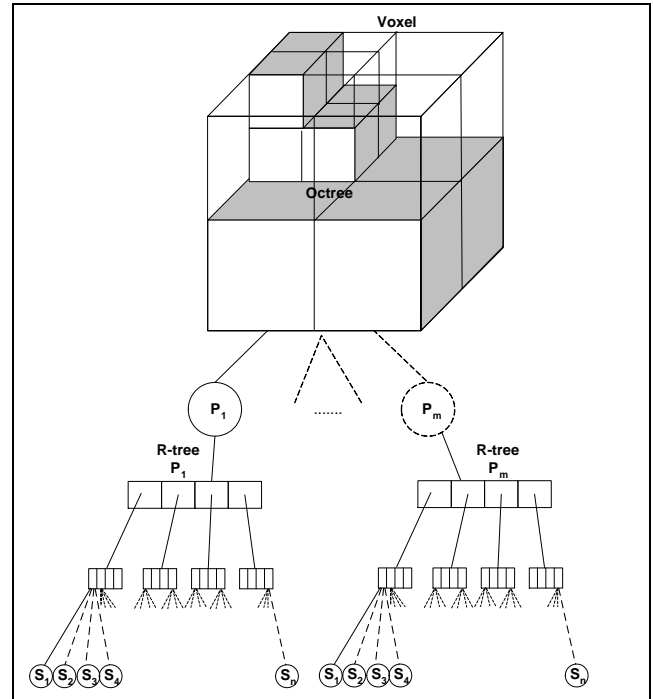

**Figure 3:** Every object of each voxel is an R-tree of points.

## 5. Bounding Sphere Hierarchy for the Collision Detection Problem

As explained in the previous section, it was decided to use a bounding sphere hierarchy like an R-tree to build the bounding volume hierarchies and organize the 3D geometry of objects for improving the performance of the collision detection process.

For the implementation of the collision detection algorithm each object is represented by an R-tree data structure in its own local coordinate system (Figure 3). Since the model has many points, first the object is partitioned using an octree. Then, a hierarchical tree is built, grouping neighboring points. The leaf nodes of the BSH point to the geometry of the partitions $P_i$ that define the object. For two objects, it checks for collisions between partitions which are in the neighborhood, eliminating comparisons with those which are far away.

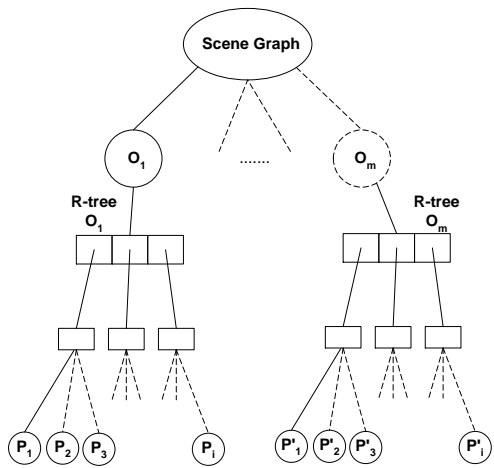

**Figure 3:** Each object of the scene graph is represented by its own sphere hieararchy data structure representing in a first level the partitions of the octree.

The same idea is applied for partitions. Partitions from a three-dimensional model can be complex with a large number of points. Another bounding spheres hierarchy is used to organize the points spatially and hence to quickly reject points that cannot intersect (Figure 4). In this approach, an R-tree is computed for each partition $P_i$, grouping neighboring points to eliminate comparisons with those that are faraway from the area of intersection. The leaf nodes of the tree point to the bounding spheres $S_i$ of the points.

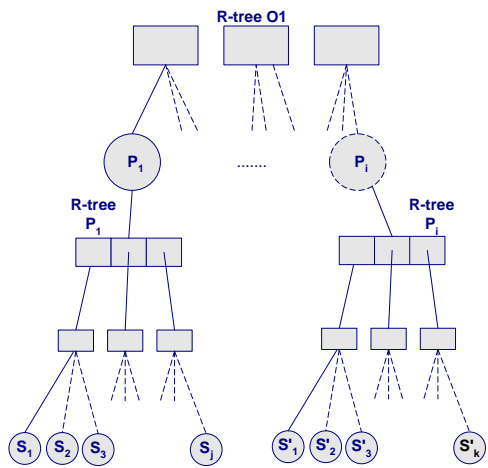

**Figure 4:** Each octree partition of the 3D model is also represented by its own bounding spheres hierarchy data structure.

## 6. COLLISION DETECTION ALGORITHM

This section presents a novel algorithm for determining intersections between pairs of 3D objects defined as point clouds. The approach presented is supported by hierarchies of bounding spheres.

The advantage of using bounding spheres is that they are good approximations to points and are faster to update.

Two levels of bounding sphere hierarchies are used. The top sphere hierarchy organizes the space divided by the octree structure to quickly reject regions of the objects that cannot intersect because they belong to partitions that are not intersecting. The lower sphere hierarchy organizes the point cloud geometry of each partition to speed up the process of finding a pair of points in close proximity to determine if the pair of objects is overlapping.

To find if two objects are intersecting, the collision detection manager makes a recursive traversal of two sphere bounding volumes trees $A$ and $B$. Figure 5 presents the pseudo code of the novel approach.

```
collidePartitions (A, B)
1:SV_B(A)=M_{B←A}•SV_A(A) //update sphere BV
2:if (SV_B(A) do not intersect SV_B(B))
3:     return
4:if A and B are leaves
5:   collidePoints (A,B)
6:else
7:   if A is an inner node and B is a leaf node then
8:     for all children A[i] do
9:       collidePartitions (A[i], B)
10:      else
11:    for all children A[i] and B[j] do
12:      collidePartitions (A[i], B[j])
```

**Figure 5:** Pseudo-code for finding pairs of partitions candidates for collision.

The collision detection algorithm first checks if the node's bounding spheres from the partition hierarchy of objects $A$ and $B$ are disjoint (line 1-2 in Figure 5).

The sphere volumes of each object are originally computed in the object's local coordinate system, $SV_A(A)$ and $SV_B(B)$, respectively. The transformation matrix that converts the local representation of object $A$ into the local coordinate system of object $B$ is defined as $M_{B←A}$. The sphere volume of object $A$ is updated to the coordinate system of object $B$, $SV_B(A)$ (line 1 of Figure 5). Once the sphere volumes of each object are in the same coordinate system they can be checked for overlap (line 2 of Figure 5).

For those pair of nodes $A$ and $B$ whose corresponding sphere volumes do not overlap, then the corresponding two objects are not intersecting and the process ends (line 3 of Figure 5).

If they overlap, they are candidate for collision, further tests must be made. If both nodes $A$ and $B$ are leaves then the sphere hierarchies of the points for this partition of the space must be checked to find a pair of points in close proximity (line 5 of Figure 5). In the case that at least one of nodes $A$ or $B$, are not leaves then it continues the traversal of the sphere bounding volume hierarchy, down to the leaves nodes (lines 6 to 12 of Figure 5).

Figure presents 6 the pseudo code for the process of finding points in close proximity for each partition.

At this point, the process of finding overlapping between the two objects uses the lower sphere hierarchy that organizes the point cloud geometry of each partition. The points of $A$ and $B$ are stored at the leaf nodes of the sphere tree. This process is very similar to the previous one, since it is a recursive traversal of two sphere bounding volumes trees *A* and *B*.

First, the sphere volume of object *A* is updated to the coordinate system of object *B*, $SV_B(A)$ (line 1 of Figure 6). If the node's bounding spheres from objects *A* and *B* are disjoint (line 2 of Figure 6) then there is no overlapping and the process ends (line 3 of Figure 6).

```
collidePoints (A, B)
1:SV_B(A) =M_{B←A}•SV_A(A) //update sphere BV
2:if (SV_B(A) do not intersect SV_B(B))
3:      return
4:if A and B are leaves
5:    return A and B are overlapping
6:else
7:   if A is an inner node and B is a leaf node then
8:     for all children A[i] do
9:        collidePoints (A[i], B)
10:       else
11:    for all children A[i] and B[j] do
12:       collidePoints (A[i], B[j])
```

**Figure 6:** Pseudo-code for finding if there is any pair of points in close proximity in a partition.

If they overlap, they are candidate for collision. If both nodes *A* and *B* are leaves then points *A* and *B* are in close proximity and the two objects are overlapping (lines 4 and 5 of Figure 6).

In the case that at least one of nodes *A* or *B*, are not leaves then it continues the traversal of the sphere bounding volume hierarchy, down to the leaves nodes (lines 6 to 12 of Figure 5).

## 7. EXPERIMENTAL RESULTS

This section presents the performance evaluation results of the novel collision detection algorithm for point cloud models described in this paper.

Point cloud models derived from laser scans present two main characteristics: they are not segmented, and points are only samples of the surface, making an actual collision between points a less likely event. Point based rendering algorithms such as QSplat [34] change the thickness and shape of a point splat to better convey visually the underlying surface while viewing in close range. Similarly we use the average closest point distance of a point divided by two to create bounding spheres at point level that ensure collision detection of the surface they represent.

As mentioned in section 3, we created an interactive system to study various collision scenarios using the Batalha Façade Model. This model was obtained using a laser scan and contains 587923 points. The avatar is the Al model using 3617 points and walks along a predefined path of 40 seconds used to benchmark our collision algorithm and depicted with a white dashed line in Figure 2.

We run our experiment on a laptop equipped with a Core 2 Duo T9300 2.50 GHz Cpu, 4 Gb of RAM memory, a NVIDIA 8800M GTX graphic card with 512 Mb and running Windows 7 64 bits. Our walkthrough application was implemented in C++ using GLUT and OpenGL libraries. The Batalha Model was partitioned using an octree data-structure with 4096 cells.

Our application was designed to run a synchronized rendering loop of 30 fps which is sufficient for desktop based real-time visualization. We should notice that the collision test is defined between the avatar and the scene points.

Table 1 presents the average frame rate and memory usage obtained with bounding sphere hierarchies of different degrees. We can notice that our approach provides an average frame rate from 15 up to 21 fps with collision detection. The application determines intersections interactively.

The higher the degree of the tree, the smaller is the height of the tree. Minimizing the height of the tree is desirable, so that when searching the hierarchy, the tree is traversed from root to a leaf in a small number of steps by visiting a fewer number of nodes. However, searching trees of a higher degree can be less efficient than searching trees of a lower degree. Trees of a higher degree need more processing at each node, since more comparisons are needed to find which leafs to descend next. Although, fewer nodes are visited during the traversal of a higher degree tree, more comparisons may be required, reducing the overall performance. Therefore a balance between the degree and height is desired.

Best results are obtained with a bounding sphere hierarchy of degree 10 for the Batalha model.

**Table 1:** Timing and memory usage for different degrees of the Bounding Sphere Hierarchy of the Batalha Model.

| Degree BSH | 4 | 6 | 8 | 10 | 12 | 14 | 16 |
|---|---|---|---|---|---|---|---|
| Avg. Fps | 18.75 | 18.19 | 17.38 | 21.25 | 15.40 | 17.88 | 16.14 |
| *Mem.(Mb)* | 39.46 | 36.84 | 35.21 | 45.95 | 45.15 | 36.50 | 45.97 |

The collision test is defined between the avatar and the scene points. Our experiments have shown that best performance is achieved using bounding sphere hierarchies of degree less than 10 (Table 1).

Figure 7 depicts the variation of the frame rate along our 40 second path using bounding sphere hierarchies of different degrees. The black dashed line corresponds to the frame rate obtained with the walkthrough of the path without collision detection. Figure 7 shows that even with collision detection the navigation is still interactive and the cost of the traversal of the hierarchies is variable due to the spatial partition of the structure.

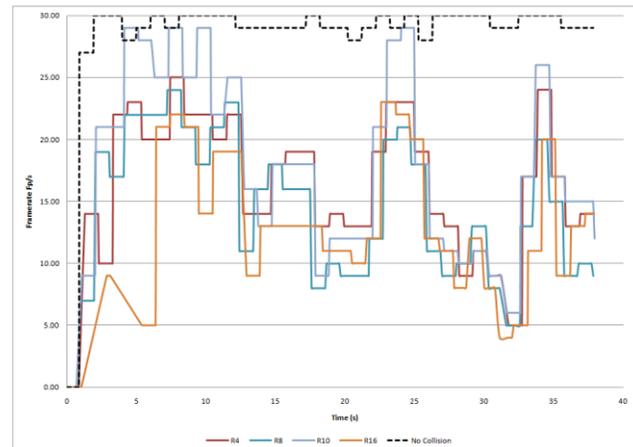

**Figure 7:** Application frame rate when Al model is walking along the path.

Figure 8 shows for the entire model the bouding spheres of the fourth level. Depending of the octree data structure the model is organized in partitions according to its point data set. In this way, depending on the density and on the octree partitions used for the model subdivision, the different bounding sphere trees do not have the same depth.

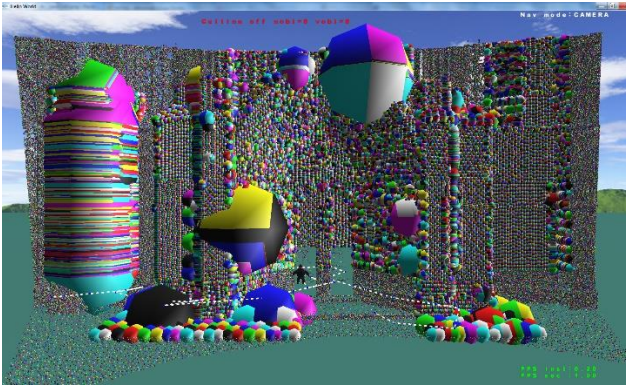

**Figure 8:** Sphere Hierarchy Batalha monastery model at level 4 for every partition.

## 8. CONCLUSIONS AND FUTURE WORK

This paper presents a novel approach for collision detection of point clouds. There are many approaches and algorithms to determine collisions between 3D polygonal models. There is very little in the literature about collision between 3D point clouds models. However, point clouds have become a popular shape representation. One of the reasons is due to the fact that 3D scanning devices became affordable and widely used for projects like VIZIR.

The proposed collision detection approach divides the scene graph in voxels. This paper implements a novel sphere tree based on an R-Tree for collision detection. There is a bounding volume R-Tree of bounding spheres for each object in a voxel that organizes spatially its point cloud.

Experimental results show that this implementation is effective in determining interactively intersections between 3D models. However, traversing tighter areas such as the door of the Monastery "still" became more difficult without colliding. In the future we would like to design an adaptive bounding box size to better handle point sets with heterogeneous sampling density. And explore an approach that manually classifies border vertices, and treats these groups of vertices as different surface entities that directly represent the geometry. We also would like to design a bounding sphere hierarchy for the partitions that distributes the point sets among different partitions in order to make the trees of the complete data model with the same heights.

## 9. ACKNOWLEDGMENTS

This work was supported by FCT (INESC-ID multiannual funding) through the PIDDAC Program funds. The authors would like to thank Instituto de Gestão do Património Arquitectónico e Arqueológico (IGESPAR) and "Artescan, 3D Scanning" for the model of the Batalha cathedral. The work presented in this paper was funded by the Portuguese Foundation for Science and Technology (FCT), VIZIR project grant (PTDC/EIA/66655/2006). In addition, Bruno Araújo would like to thank FCT for doctoral grant reference SFRH/ BD/ 31020/ 2006.